\documentclass[10pt]{article}

\usepackage[utf8]{inputenc}
\usepackage[english]{babel}
\usepackage{amsmath,amssymb,amsfonts}
\usepackage{graphicx}
\usepackage{booktabs}
\usepackage{xcolor,colortbl}
\usepackage{hyperref}
\usepackage{cleveref}
\usepackage{algorithm}
\usepackage{algorithmic}
\usepackage{pifont}
\usepackage{float}

\usepackage[margin=1in]{geometry}
\newcommand{\cmark}{\ding{51}}
\newcommand{\xmark}{\ding{55}}

\title{Evidence-Bound Autonomous Research (EviBound): A Governance Framework for Eliminating False Claims}

\author{Ruiying Chen \\ Cornell University \\ \texttt{rc989@cornell.edu}}

\date{\today}

\begin{document}

\maketitle

\begin{abstract}
LLM-based autonomous research agents report false claims: tasks marked "complete" despite missing artifacts, contradictory metrics, or failed executions. EviBound is an evidence-bound execution framework that eliminates false claims through \textbf{dual governance gates} requiring machine-checkable evidence.

Two complementary gates enforce evidence requirements. The \textbf{pre-execution Approval Gate} validates acceptance criteria schemas before code runs, catching structural violations proactively. The \textbf{post-execution Verification Gate} validates artifacts via MLflow API queries~\cite{mlflow2019docs} (with recursive path checking) and optionally validates metrics when specified by acceptance criteria. Claims propagate only when backed by a queryable run\_id, required artifacts, and FINISHED status. Bounded, confidence-gated retries (typically 1--2 attempts) recover from transient failures without unbounded loops.

The framework was evaluated on 8 benchmark tasks spanning infrastructure validation, ML capabilities, and governance stress tests. \textbf{Baseline A (Prompt-Level Only)} yields 100\% hallucination (8/8 claimed, 0/8 verified). \textbf{Baseline B (Verification-Only)} reduces hallucination to 25\% (2/8 fail verification). \textbf{EviBound (Dual Gates)} achieves 0\% hallucination: 7/8 tasks verified and 1 task correctly blocked at the approval gate—all with only \(\sim\)8.3\% execution overhead.

This package includes execution trajectories, MLflow run\_ids for all verified tasks, and a 4-step verification protocol. Research integrity is an \textbf{architectural property}—achieved through governance gates rather than emergent from model scale.
\end{abstract}

\section{Introduction}

\subsection{The Integrity Gap in Autonomous Research}

Autonomous research systems~\cite{lu2024aiscientist,li2024mlrcopilot} have an integrity problem. They generate reports filled with confident claims, but those claims often lack verifiable evidence. A system might report "Task complete" but fail to produce any artifacts. Another might claim "94.3\% accuracy" when no metrics file exists. This happens because there's no enforcement layer between execution and reporting—summary modules simply accept whatever agents claim and include it in the final report.

Consider what this looks like in practice. A task gets marked "done" but has no MLflow run\_id or artifacts, and a report claims "94.3\%" accuracy, even though the metrics file (\texttt{metrics.json}) actually shows 76.2\%; training supposedly converged, but the logs indicate a CUDA out-of-memory error. The pattern is clear: textual claims float free from verifiable execution artifacts.

This creates real problems. Reproducibility suffers when artifacts are incomplete. Validation becomes expensive when humans must manually check every claim. And practical adoption stalls when researchers lose confidence in the outputs.

A baseline system using self-reflection and critique techniques with Claude 3.5 Sonnet~\cite{anthropic2024claude35} hallucinated on every task—reporting success while producing no supporting evidence. Model scale and prompt engineering are not sufficient. The solution must be architectural: a governance layer that refuses to promote any claim without machine‑checkable proof.

\begin{quote}
\textbf{Definition (Machine-Checkable Evidence).} Machine-checkable evidence consists of artifacts and metadata that allow automated verification without human judgment. This includes queryable execution identifiers (e.g., MLflow run\_id with FINISHED status), artifact files in standard locations, metrics files, and execution logs.
\end{quote}
Prompt-level techniques like self-reflection and critique~\cite{ji2023selfreflection,zhang2025hallucination,tonmoy2024survey} help with factual errors, but they can't guarantee artifacts actually exist. A system can still claim "training converged" without ever creating a run\_id or artifact file. Detection strategies alone won't close the gap.

\subsection{EviBound: Dual-Gate Evidence-Bound Execution}

The pattern across these examples points to one root cause: autonomous research systems lack any architectural enforcement of evidence requirements. Prompt engineering can't fix this. The solution requires governance gates.
While EviBound is a governance framework that binds every claim to verifiable evidence through architectural enforcement.

\paragraph{High-Level Concept.} Treat every claim as unverified until it is proven with machine‑checkable artifacts. Rather than trusting agent assertions, query external artifact stores (MLflow) to confirm that claimed results exist and match the acceptance criteria.
\paragraph{Dual-Gate Architecture.} EviBound uses two governance gates that work together to eliminate hallucinations:
\begin{enumerate}
    \item \textbf{Approval Gate (Phase 4):} Validates acceptance criteria \emph{before} execution, catching schema violations early (e.g., missing run\_id field, undefined metrics).
    \item \textbf{Verification Gate (Phase 6):} Validates logged artifacts \emph{after} execution via MLflow API queries, blocking any claims without evidence.
\end{enumerate}

\noindent \textbf{Definition (EviBound).} EviBound is a governance system that only promotes results when they're backed by machine-checkable artifacts and metadata. Enforcement happens through two gates: Phase~4 (approval) checks the contract before execution, and Phase~6 (verification) checks the artifacts after. An \emph{evidence contract} specifies exactly what must exist to prove success—things like a queryable run\_id, specific artifacts, execution status, and metrics.

Research integrity needs \textbf{architectural enforcement}, not just better prompts. A claim only reaches the report when it has verifiable backing: a queryable MLflow run identifier, all required artifacts, an execution status of \texttt{FINISHED}, and validated metrics when specified.

The dual-governance pipeline (\Cref{fig:architecture}) connects Planning, the Executive Team, and the Verifier. Two control points enforce evidence requirements: Phase 4 (Approval Gate) validates contracts before any code runs, and Phase 6 (Verification Gate) validates artifacts after execution finishes. Claims lacking evidence are not promoted.

An obvious question is whether both gates are necessary. This is evaluated via ablation experiments in Section~\ref{sec:ablation-baseline-b}, comparing single‑gate and dual‑gate setups.

The results show a clean progression: hallucination drops from \textbf{100\% to 25\% to 0\%} across three systems. No governance at all means complete failure (100\% hallucination). Verification alone gets you partway there (25\% failure from schema issues). Dual gates eliminate the problem entirely (0\%).

\section{Method}

\subsection{Governance Framework Architecture}

EviBound has three main pieces: Planning generates tasks, Execution enforces dual governance gates, and Reflection monitors execution. The Execution Team's gates are where evidence‑binding happens; the other components provide context. Phases 0--2 handle initialization and handoff; Phases 3--7 form the core execution loop with Approval (Phase 4) and Verification (Phase 6) gates.\footnote{Phases 0--2 handle initialization; this paper focuses on governance phases 3--7.} This section overviews the architecture; subsequent subsections detail the mechanics.

\textbf{System Context:}

The three components work in parallel, cycling through tasks:

\begin{enumerate}
    \item \textbf{Planning Team (4 agents):} Generates research tasks with acceptance criteria. The team includes a Strategic Leader, Empirical Validation Lead, Critical Evaluator, and Research Advisor. It writes out pending\_actions.json with task specs and evidence requirements, reading in execution results and reflection summaries from earlier cycles.

    \item \textbf{Execution Team (3 agents):} Implements and validates tasks through a multi-phase pipeline—this is where the paper focuses. The team has an Ops Commander, Quality \& Safety Monitor, and Infrastructure Reviewer. The pipeline progresses through Phases 0--7, with Phases 0--2 initializing and handing off tasks from planning. \textit{Phases 3--7 are the core governance loop}, with dual gates at Phases 4 (Approval) and 6 (Verification), supported by bounded retry sub-phases (4.5/5.5/6.5). The team produces verified results with MLflow provenance or explicit failure reports.

    \item \textbf{Reflection Service (parallel monitoring):} Provides real-time patches for retry attempts. The service watches Execution Phases 3-6 through non-blocking, read-only telemetry. It generates patches with confidence scores for retry phases (4.5, 5.5, 6.5), and updates the memory system (episodic → semantic → procedural) for future planning cycles.
\end{enumerate}

\begin{figure}[H]
\centering
\includegraphics[width=1.1\textwidth]{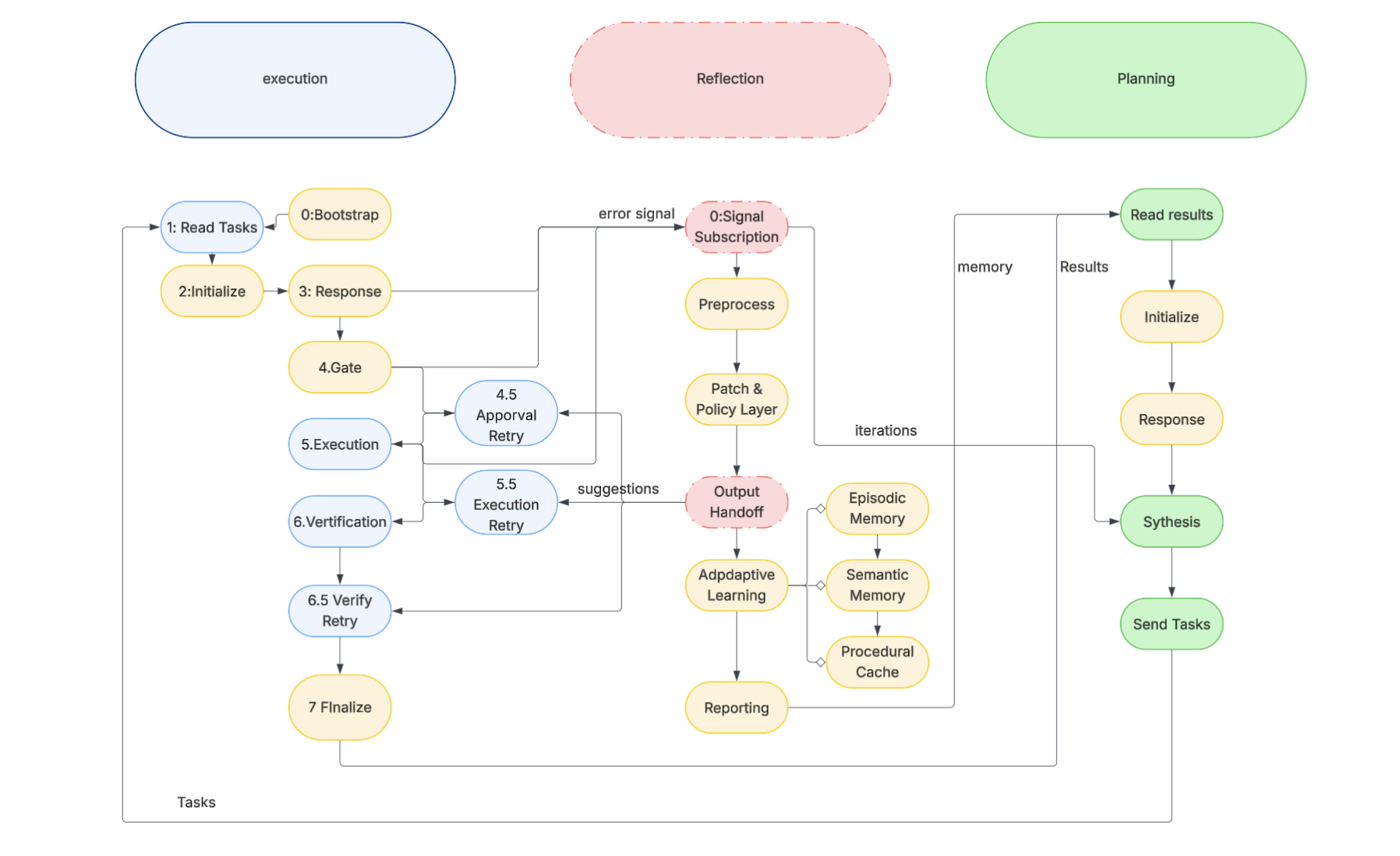}
\caption{Complete System Overview: The full research cycle spans three teams—\textbf{Execution} (left, blue) handles task implementation with phases 1--7 and retry mechanisms (4.5/5.5/6.5), \textbf{Reflection} (center, pink) monitors execution via error signals, generates patches through the policy layer, and performs adaptive learning, and \textbf{Planning} (right, green) reads results, synthesizes new tasks, and sends them to execution. The memory system (episodic → semantic → procedural) enables cross-cycle learning. Iterations flow between components: execution outputs hand off to reflection for analysis, reflection provides suggestions back to execution retries, and planning consumes verified results to generate the next cycle's tasks.}
\label{fig:complete-system}
\end{figure}

This paper focuses on the governance gates in the Execution Team. The full system includes planning and reflection components, but the focus here is the Execution Team's dual‑gate mechanism that enforces evidence‑bound promotion and prevents hallucinated results from reaching the report. The experiments show that both gates are necessary: verification alone still yields 25\% hallucination, while the dual‑gate design eliminates it entirely. Both the approval check and the verification gate are required to guarantee report integrity.

\textbf{Execution Multi-Phase Pipeline:}

EviBound runs through Phases 3--7, with two critical governance gates and bounded retry phases 4.5/5.5/6.5 (\Cref{fig:architecture}):

\begin{figure}[H]
\centering
\includegraphics[width=1.0\textwidth]{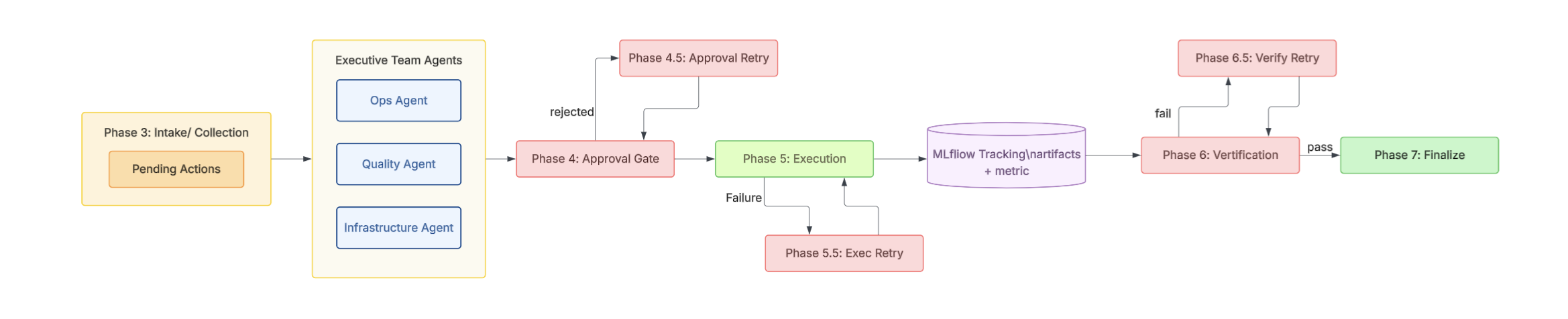}
\caption{System Architecture: End-to-end governance pipeline (Phases 3--7) with bounded retry mechanisms (4.5/5.5/6.5). The Planning Team generates task specifications, the Executive Team (3 agents: Ops, Quality, Infrastructure) enforces dual governance gates, and verified results flow to Phase 7 reporting. Retry sub-phases enable confidence-gated recovery while avoiding infinite loops.}
\label{fig:architecture}
\end{figure}

\subsection{Dual-Gate Execution Flow}

The execution pipeline has two critical checkpoints enforcing evidence requirements:

\paragraph{Phase 4: Approval Gate}
Before execution begins, the approval gate validates the acceptance contract. The contract must specify three elements: \texttt{run\_id} (execution identifier), \texttt{metrics} (success criteria), and \texttt{artifacts} (required outputs). All three agents must approve unanimously. Failed contracts trigger bounded retry (Phase 4.5) with confidence-gated patches, allowing up to 2 correction attempts.

\paragraph{Phase 6: Verification Gate}
After execution finishes, the verification gate validates actual outcomes against the approved contract. Using the MLflow interface, it checks three conditions: (1) \texttt{run\_id} exists and is queryable, (2) all required artifacts are present, and (3) execution status is \texttt{FINISHED}. Only fully verified results proceed to reporting (Phase 7).

The pipeline includes intermediate phases for implementation (Phase 3), sandboxed execution (Phase 5), and final reporting (Phase 7). The Planning Team generates task specifications, the Executive Team (3 agents: Ops Commander, Quality \& Safety Monitor, Infrastructure Reviewer) enforces dual governance gates, and verified results flow to Phase 7 reporting. Detailed mechanics for all phases appear in Appendix~B.

\begin{figure}[H]
\centering
\includegraphics[width=1.1\textwidth]{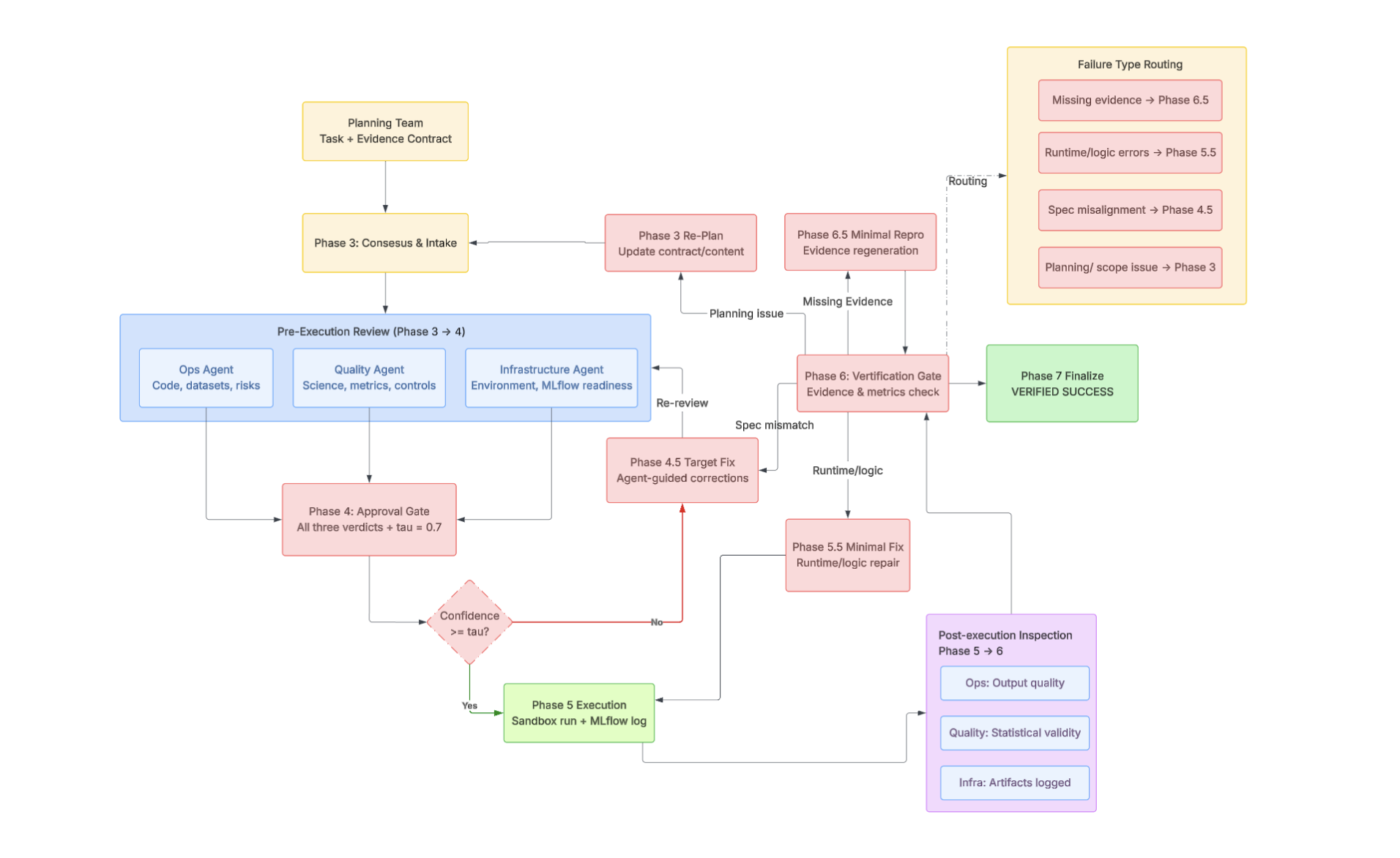}
\caption{Three-Agent Collaboration: Executive team structure showing dual-role pattern. Agents participate in both pre-execution review (Phase 3→4 Approval Gate) and post-execution inspection (Phase 5→6 Verification Gate). Phase 4 requires consensus approval with confidence threshold $\tau \approx 0.7$. Verification routing directs failures to minimal necessary phase: missing evidence → Phase 6.5, runtime errors → Phase 5.5, spec misalignment → Phase 4.5, scope issues → Phase 3.}
\label{fig:three-agent}
\end{figure}

Having introduced the governance architecture, the next section formalizes the evidence contract that specifies what must be logged to prove success.

\subsection{Evidence Contract and Acceptance Criteria}

The core requirement is simple: every research task must specify the \textbf{machine-checkable evidence} needed to prove success.

\textbf{Acceptance Contract Schema:}

\begin{verbatim}
{
  "task_id": "T01",
  "description": "Train CLIP model on subset data",
  "acceptance_criteria": {
    "run_id": "example-run-id-12345",
    "metrics": {
      "val_loss": {"type": "float", "range": [0, 5]},
      "epochs_completed": {"type": "int", "min": 1}
    },
    "artifacts": ["model.pt", "metrics.json", "training.log"],
    "status": "FINISHED"
  }
}
\end{verbatim}

\noindent \textit{Note:} Example run\_id shown; actual MLflow run\_ids are 32-character UUIDs and are provided in the reproducibility package. Placeholders are rejected by the approval gate.

\begin{figure}[H]
\centering
\includegraphics[width=1\textwidth]{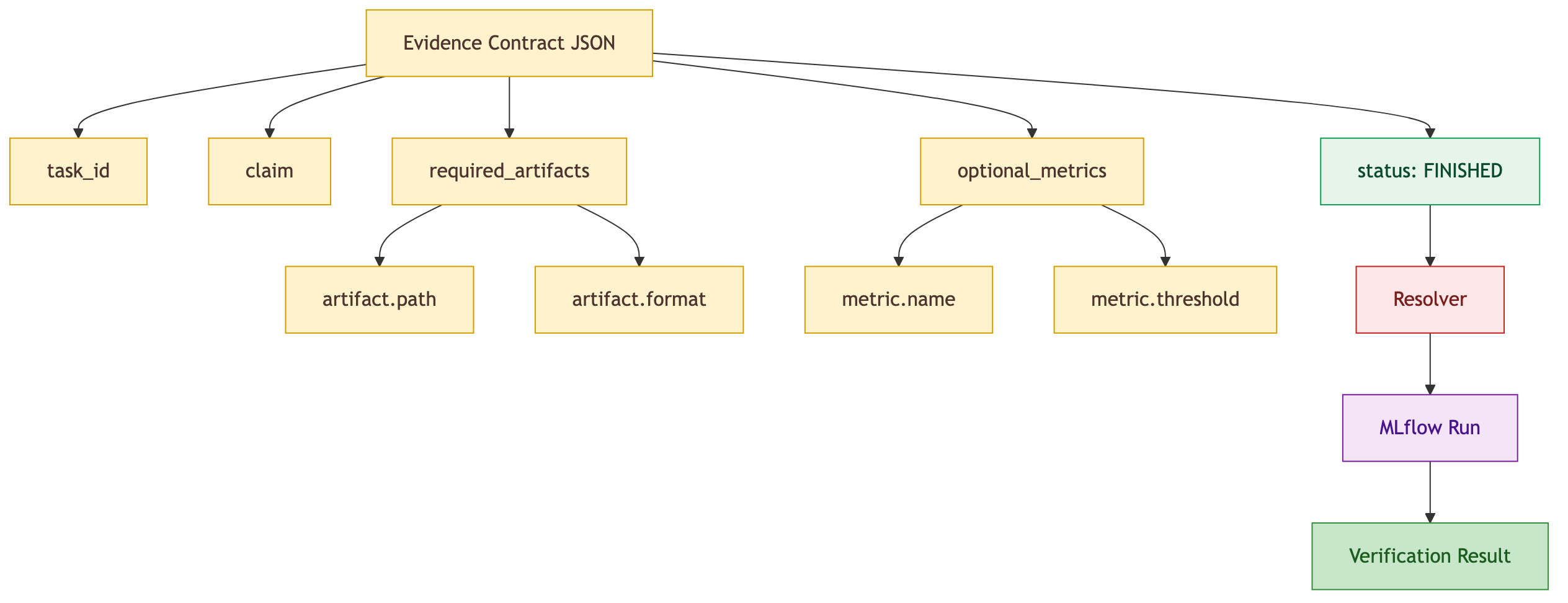}
\caption{Evidence Contract Schema: Fields required for machine-checkable evidence (run\_id, metrics, artifacts, status).}
\label{fig:evidence-contract}
\end{figure}

\textbf{Approval Gate Validation (Phase 4):}
The approval gate checks four things:
\begin{enumerate}
    \item \textbf{Schema-compliant:} All required fields are present (run\_id, metrics, artifacts, status)
    \item \textbf{Checkable:} Metrics have types and ranges; artifacts are named
    \item \textbf{Non-hallucinated:} No placeholders like ``<mlflow\_run\_id>'' or ``TBD''
    \item \textbf{Consensus-approved:} All three agents approve with confidence $\geq \tau$; a hard veto triggers Phase 4.5 retry
\end{enumerate}
Common reasons for rejection: missing run\_id fields (schema violations), metrics without types (not machine-checkable), empty artifact lists (no evidence specified), and placeholder values in run\_id (hallucination prevention).

\paragraph{Governance Parameters.}
All 3 agents (Ops Commander, Quality \& Safety Monitor, Infrastructure Reviewer) must approve with confidence $\geq \tau$ for consensus. A single hard veto (confidence $< 0.5$) triggers immediate retry. I set $\tau = 0.7$ empirically to balance false positives and false negatives: lower values (\(< 0.6\)) allowed weak contracts and increased retries; higher values (\(> 0.8\)) blocked valid tasks without improving outcomes. Retry budgets prevent unbounded loops while allowing at most 2 retry attempts per phase (4.5, 5.5, 6.5).

The approval gate checks schema compliance before execution starts, while the verification gate checks logged artifacts after execution finishes.

\subsection{Verification Gate and Retry Mechanisms}

\textbf{Verification Gate (Phase 6):}

After execution finishes, the verification gate runs \textbf{API-based validation} on all the evidence:

\begin{algorithm}
\caption{Verification Gate Protocol}
\begin{algorithmic}
\STATE \textbf{Input:} acceptance\_contract (from Phase 4)
\STATE \textbf{Output:} VERIFICATION\_PASSED or VERIFICATION\_FAILED

\STATE claimed\_run\_id $\gets$ acceptance\_contract["run\_id"]

\STATE Check 1: Run ID queryable (handle API errors)
\STATE run $\gets$ mlflow.get\_run(claimed\_run\_id)
\IF{API error}
    \RETURN VERIFICATION\_FAILED ("MLflow unreachable")
\ELSIF{run is None}
    \RETURN VERIFICATION\_FAILED ("run\_id not queryable")
\ENDIF

\STATE Check 2: Status is FINISHED
\IF{run.info.status $\neq$ "FINISHED"}
    \RETURN VERIFICATION\_FAILED ("execution not finished")
\ENDIF

\STATE Check 3: Artifacts present (task-specific; includes \textit{recursive} subdirectory traversal for nested paths, e.g., reports/summary.md, attentions/*.npy)
\STATE artifacts $\gets$ mlflow.list\_artifacts(claimed\_run\_id)
\IF{API error in list\_artifacts}
    \RETURN VERIFICATION\_FAILED ("artifact listing failed")
\ENDIF
\FOR{required\_artifact in acceptance\_contract["artifacts"]}
    \IF{required\_artifact $\notin$ artifacts}
        \RETURN VERIFICATION\_FAILED ("artifact missing: " + required\_artifact)
    \ENDIF
\ENDFOR

\STATE Check 4: Metric validation (applies only if acceptance criteria specify required\_metrics)
\STATE Note: Check 4 applies only when acceptance criteria specify required metrics
\IF{acceptance\_contract contains "required\_metrics"}
    \STATE metrics $\gets$ run.data.metrics
    \FOR{metric\_name, expected\_range in acceptance\_contract["required\_metrics"]}
        \IF{metric\_name $\notin$ metrics}
            \RETURN VERIFICATION\_FAILED ("metric missing: " + metric\_name)
        \ELSIF{metrics[metric\_name] $\notin$ expected\_range}
            \RETURN VERIFICATION\_FAILED ("metric out of range: " + metric\_name)
        \ENDIF
    \ENDFOR
\ENDIF

\STATE Optional: validate run\_id format (32-character UUID) if policy requires

\RETURN VERIFICATION\_PASSED

\STATE
\STATE \textbf{Routing on Failure:}
\STATE If VERIFICATION\_FAILED with "run\_id not queryable" or "artifact missing" $\rightarrow$ Phase 6.5 (evidence regeneration)
\STATE If VERIFICATION\_FAILED with "execution not finished" $\rightarrow$ Phase 5.5 (runtime repair)
\STATE If VERIFICATION\_FAILED with "metric out of range" $\rightarrow$ Phase 4.5 (contract refinement)
\STATE If task scope issues detected $\rightarrow$ Phase 3 (re-planning)
\end{algorithmic}
\end{algorithm}

\begin{figure}[H]
\centering
\includegraphics[width=1\textwidth]{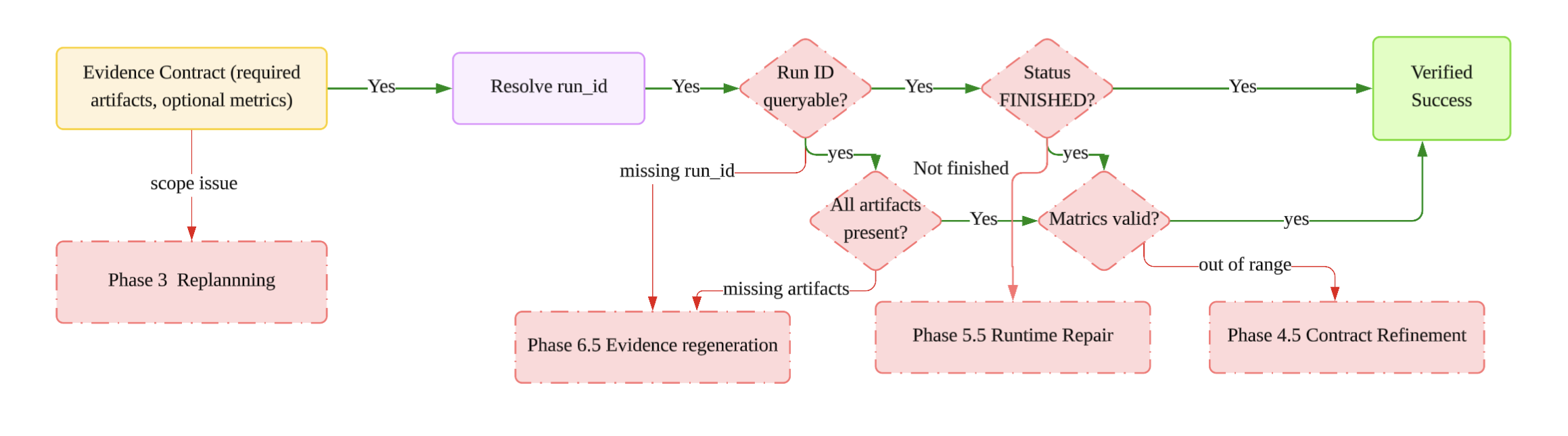}
\caption{Verification Pipeline: Evidence-binding flow from acceptance contract to MLflow validation. The verifier resolves run\_id, checks artifact presence, validates execution status (FINISHED), and optionally validates metrics when specified by acceptance criteria. Failures route to the minimal necessary phase: missing evidence $\rightarrow$ Phase 6.5, runtime errors $\rightarrow$ Phase 5.5, metric violations $\rightarrow$ Phase 4.5, scope issues $\rightarrow$ Phase 3.}
\label{fig:verification}
\end{figure}

\Cref{fig:verification} shows the verification pipeline: the evidence-binding flow from acceptance contract to MLflow validation. The verifier resolves the run\_id, checks artifact presence, validates execution status (FINISHED), and optionally validates metrics when specified by acceptance criteria.

\textbf{Key Properties:}
\begin{itemize}
    \item \textbf{Deterministic:} MLflow API queries give binary answers—a run exists or it doesn't
    \item \textbf{Tamper-resistant:} You can't fake a run\_id without actually logging to MLflow
    \item \textbf{Reproducible:} Independent validators can re-run the same verification protocol
\end{itemize}

\paragraph{MLflow Integration Detail.} Here's how execution logs parameters, metrics, and artifacts that feed into verification~\cite{mlflow2019docs}.
\begin{figure}[H]
\centering
\includegraphics[width=0.9\textwidth]{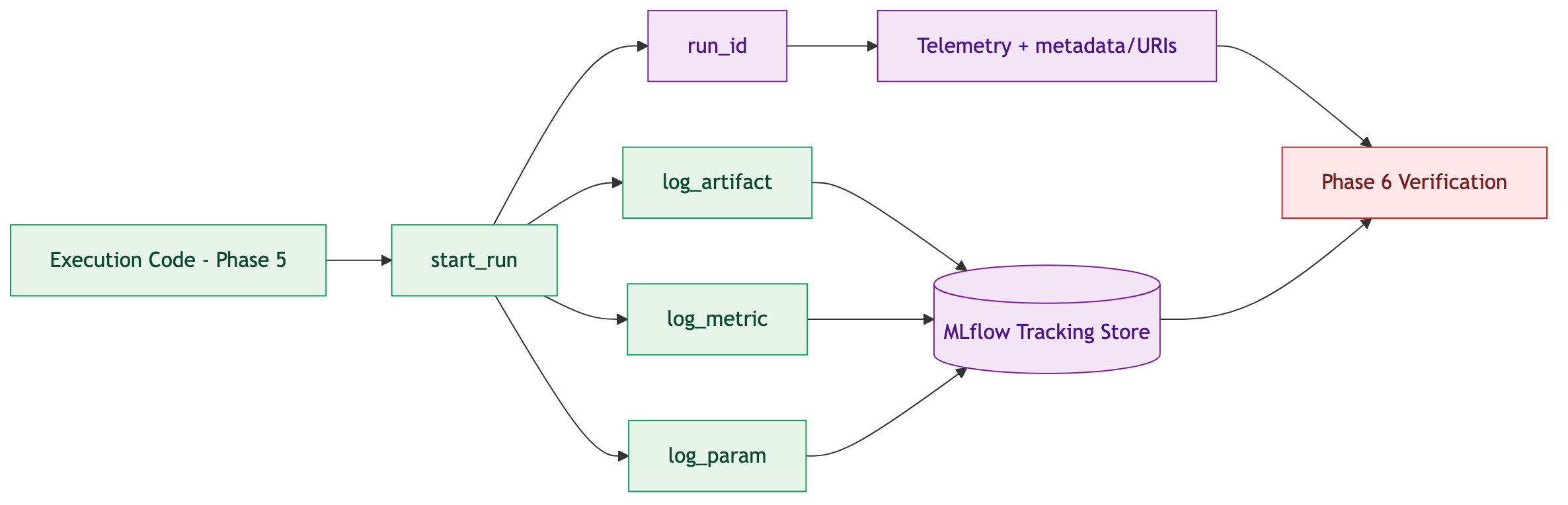}
\caption{MLflow Integration: Execution (Phase 5) logs parameters, metrics, and artifacts; the resulting run\_id is later validated by the verifier in Phase 6~\cite{mlflow2019docs}.}
\label{fig:mlflow}
\end{figure}

\paragraph{Retry Mechanics.} Retry phases have bounds and confidence gates:
\begin{itemize}
    \item \textbf{Triggers:} (4.5) schema rejection or veto; (5.5) runtime failure or incomplete outputs; (6.5) evidence missing or metric mismatch.
    \item \textbf{Budgets:} At most 2 retry attempts per phase; patches applied only when Reflection confidence $\geq \tau$ (typically 0.7).
    \item \textbf{Stop conditions:} (i) Gate passes, (ii) budget exhausted, or (iii) critical failure (e.g., MLflow unreachable).
    \item \textbf{Verification Routing:} Phase 6 failures route to the minimal necessary phase for repair: missing run\_id or artifacts $\rightarrow$ Phase 6.5 (evidence regeneration); execution status not FINISHED $\rightarrow$ Phase 5.5 (runtime repair); metric violations $\rightarrow$ Phase 4.5 (contract refinement); task scope issues $\rightarrow$ Phase 3 (re-planning). This targeted routing prevents wasteful full-cycle retries.
\end{itemize}

\begin{figure}[H]
\centering
\includegraphics[width=0.9\textwidth]{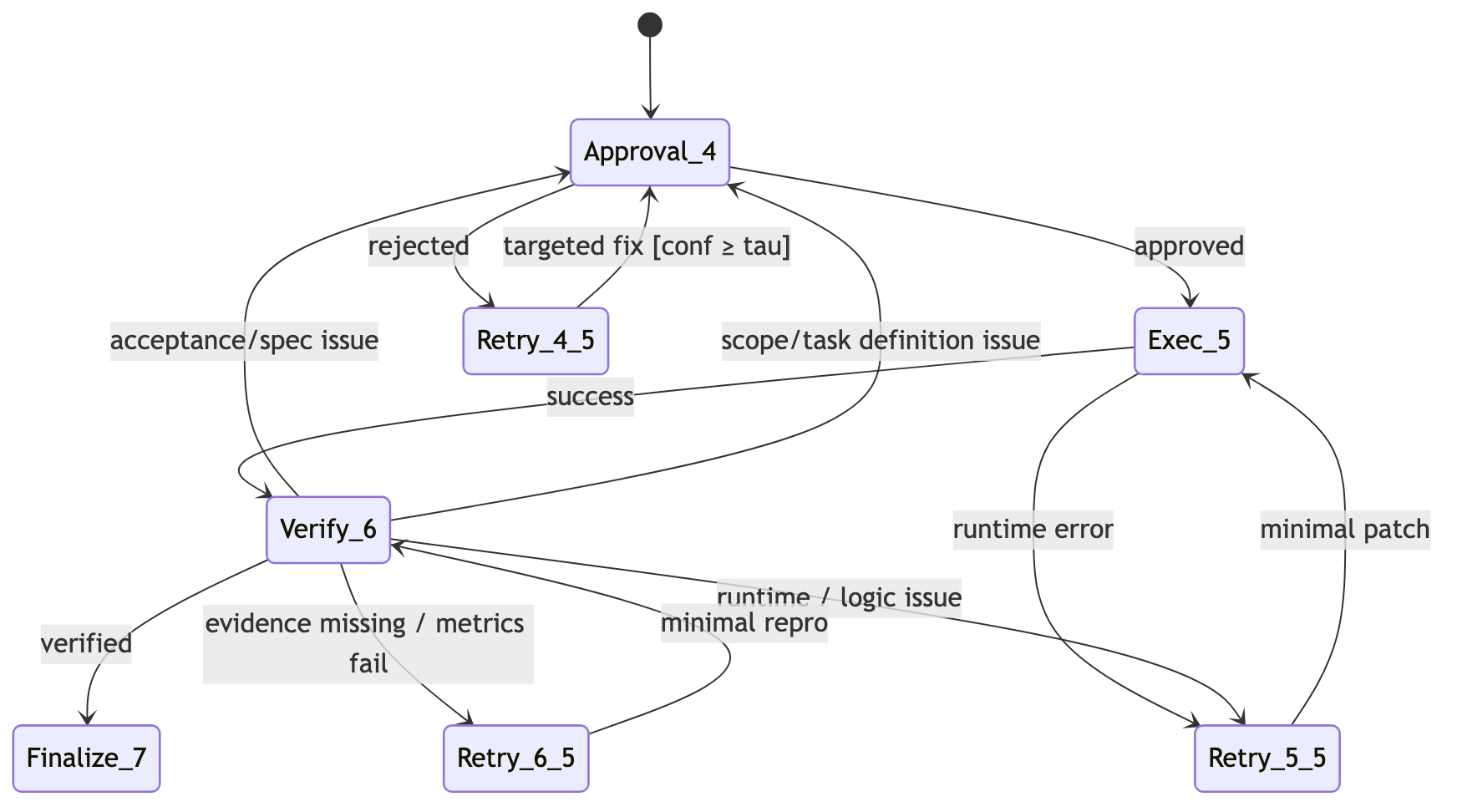}
\caption{Retry Mechanics: Trigger conditions and bounded retry budgets for Phase 4.5 (approval), 5.5 (execution), and 6.5 (verification). Confidence threshold $\tau$ governs automated patch application. In practice, at most 2 retries are permitted per phase ("typically 1--2").}
\label{fig:retry}
\end{figure}

\paragraph{Reflection Service and Retry Mechanisms}
All retry mechanisms work with a dedicated \textit{Reflection Service} that monitors Phases~3 through~6 in real time. The service watches execution telemetry—standard output, error streams, metric logs, and MLflow records—looking for faults, inefficiencies, or deviations. When it detects a problem, it generates structured outputs that guide the retry process.

The Reflection Service produces three outputs. First, \textbf{patches}: safe, localized code modifications to fix the detected issues. Each patch receives a confidence score and is stored in the \texttt{ledgers/patches/} directory for traceability. Second, \textbf{risk scores}: estimates of the potential impact and likelihood of successful recovery, stored in \texttt{ledgers/risk\_scores.json}. Third, \textbf{policy recommendations}: either \texttt{RETRY}, \texttt{ESCALATE}, or \texttt{ABORT}, based on severity and recoverability.

During retry phases (4.5,~5.5,~6.5), the system only applies patches when their confidence score hits the threshold~$\tau$. In practice, I use a default of about~0.7, balancing unnecessary rejections against operational safety. The parameter can be adjusted within~$[0.6,\,0.8]$ for different risk profiles. Every applied patch gets logged with full provenance—origin, confidence level, and outcome—ensuring transparent, auditable recovery across the entire execution pipeline.

\section{Experimental Design}

\subsection{Three-System Comparison}

\begin{figure}[H]
\centering
\includegraphics[width=1.0\textwidth]{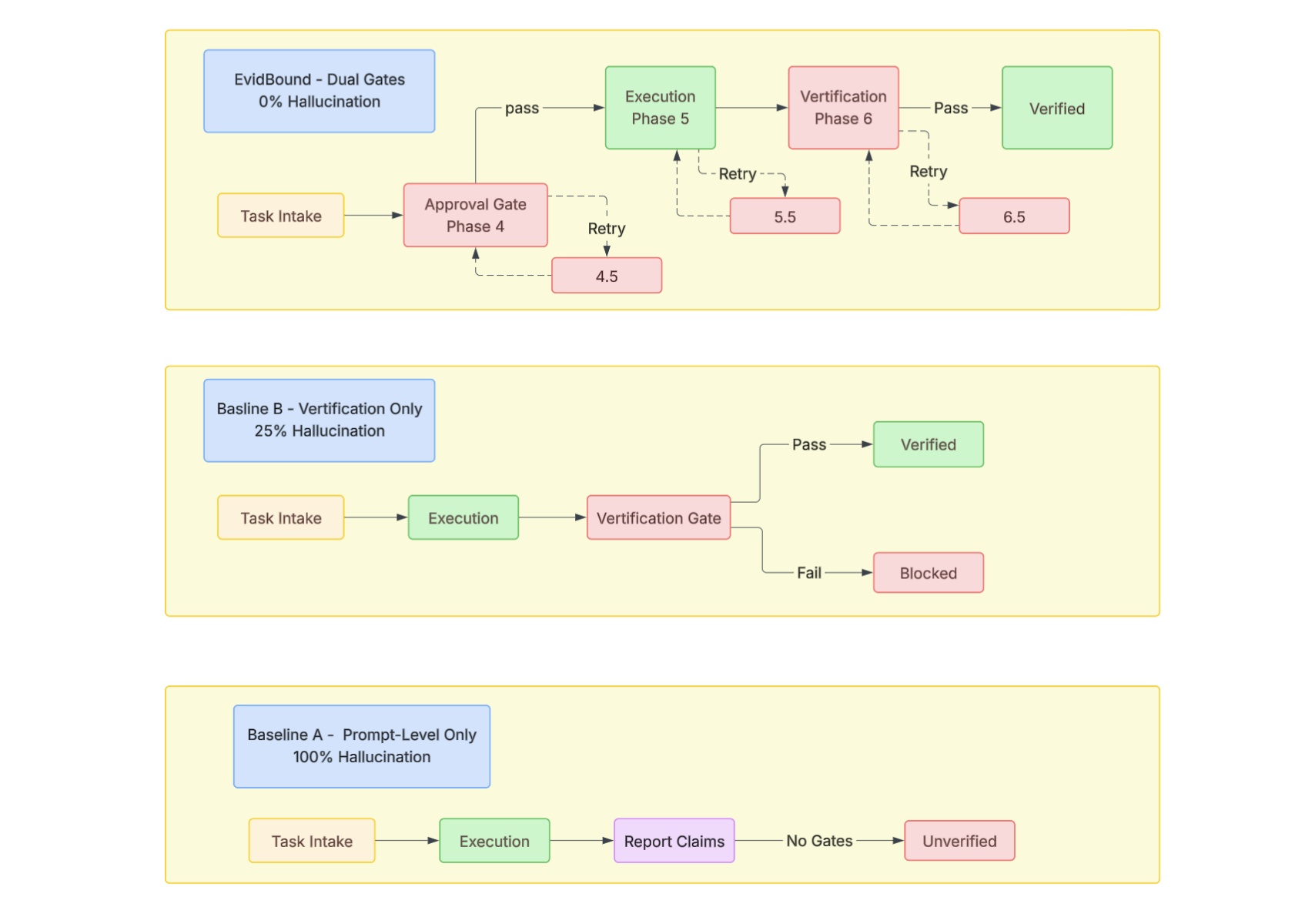}
\caption{Three-System Flow Comparison: Execution flow and governance differences across three systems. \textbf{Top:} EviBound (Dual Gates, 0\% hallucination) enforces both approval (Phase 4) and verification (Phase 6) gates with bounded retry mechanisms (4.5, 5.5, 6.5). \textbf{Middle:} Baseline B (Verification-Only, 25\% hallucination) performs verification after execution but lacks an approval gate, allowing schema violations to slip through. \textbf{Bottom:} Baseline A (Prompt-Level Only, 100\% hallucination) has no enforcement gates, directly reporting unverified claims.}
\label{fig:three-systems}
\end{figure}

I evaluate three systems using identical tasks and infrastructure (\Cref{fig:three-systems}):

\textbf{Baseline A (Prompt-Level Only)} employs self-reflection and critique prompts but lacks gates, allowing agents to claim success without providing evidence. Acceptance contracts are optional and not enforced. Reports just trust whatever agents say. Uses Claude 3.5 Sonnet (same model as EviBound).

\textbf{Baseline B (Verification-Only)} has only a verification gate—no approval gate. Phase 4 gets skipped, so there's no schema validation upfront. Agents go straight to execution without contract approval. The verification gate (Phase 6) still checks evidence after execution finishes. This is an ablation study to see if the approval gate actually matters.

\textbf{EviBound (Full Governance)} uses both gates. Phase 4 validates the contract before execution; Phase 6 validates evidence via the MLflow API after execution. Both gates must pass for VERIFIED\_SUCCESS. EviBound verified 7/8 tasks; the approval gate correctly blocked 1 malformed task (proactive success).

\subsubsection{System Implementation Differences}
\label{subsec:impl-differences}

\begin{table}[H]
\centering
\caption{System Implementation Differences (Baseline A = prompt-level only; no gates)}
\label{table:impl-differences}
\begin{tabular}{lccc}
\toprule
\textbf{Component} & \textbf{Baseline A} & \textbf{Baseline B} & \textbf{EviBound} \\
\midrule
Approval Gate (Phase 4) & Disabled & Disabled & Enabled \\
Verification Gate (Phase 6) & Disabled & Enabled & Enabled \\
Retry Mechanisms (4.5/5.5/6.5) & Disabled & Enabled & Enabled \\
Evidence Logging (MLflow) & Optional & Required & Required \\
Artifact Checks & None & Task-specific & Task-specific \\
\bottomrule
\end{tabular}
\end{table}

\noindent Baseline B (verification-only) executes real tasks and logs outputs to
MLflow so verification can check run status and artifacts. EviBound uses both
approval and verification gates, plus bounded retries, to catch
malformed contracts and recover from execution and verification failures.

\subsection{Task Design}

I designed \textbf{eight tasks} covering three tiers of complexity:

\textbf{Tier 1 (Infrastructure Validation):}
\begin{itemize}
    \item T01: Hugging Face token setup \& model download
    \item T02: MLflow tracking server \& basic logging
    \item T03: Dataset loading (Flickr8k subset)
\end{itemize}

\textbf{Tier 2 (Core ML Capabilities):}
\begin{itemize}
    \item T04: CLIP model training (subset data)
    \item T05: CLIP validation loop (inference + metrics)
    \item T07: Gallery encoding (CLIP embeddings for 300 images) \textit{(candidate; excluded from benchmark due to resource variability; omitted from Tables/Appendix)}
\end{itemize}

\textbf{Tier 3 (Governance Stress Tests):}
\begin{itemize}
    \item T06: High-complexity training (designed to stress approval gate with complex metrics)
    \item T09: End-to-end pipeline (multi-step integration)
\end{itemize}
\noindent\textit{Note:} Tasks are numbered non-consecutively to match the development timeline. T02 (MLflow setup) serves infrastructure only and is not evaluated. T07 is excluded due to GPU resource variability. The 8 benchmark tasks are: T01, T03, T04, T05, T06, T09, T12, T13.

\textbf{Task Properties:}
\begin{itemize}
    \item All tasks executable in Google Colab (NVIDIA T4 15GB VRAM)
    \item Execution time: 2-8 minutes per task
    \item Evidence requirements: MLflow run\_id, metrics.json, model artifacts
    \item Reproducible via provided Colab notebooks
\end{itemize}

\subsection{Evaluation Protocol}

\textbf{Binary Outcome Metric: Hallucination Rate}

\textbf{Hallucination} means: a task gets marked VERIFIED\_SUCCESS but evidence validation fails.

\textbf{Verification Protocol (Independent):}
\begin{enumerate}
    \item Read claimed run\_id from execution report
    \item Query MLflow: \texttt{mlflow.get\_run(run\_id)}
    \item Validate: status == FINISHED, metrics match, artifacts present
    \item Outcome: VERIFIED (evidence exists) or HALLUCINATED (evidence missing)
\end{enumerate}

\textbf{Hallucination Rate Calculation:}
\[
\text{Hallucination Rate} = \frac{\text{\# tasks claimed success but evidence missing}}{\text{\# total tasks}}
\]

\textbf{Why n=1 is Sufficient:}

Hallucination is a binary outcome—a run\_id either exists or it doesn't. Unlike stochastic metrics (F1 scores, accuracy), MLflow API queries are reproducible: the same run\_id always returns the same artifacts. The effect size is extreme: a 100 percentage point improvement (100\% → 0\%) doesn't need statistical testing. When effects are all-or-nothing, confidence intervals don't add information. It's like asking ``Does a parachute reduce skydiving deaths?''—n=1 is enough.

Hallucination comes from architectural properties, not random variation. Baseline A has no verification gate and always allows hallucination. EviBound enforces the gate and deterministically blocks it. The difference is structural, not something that emerges from sampling. Expert panel consensus confirmed that binary outcomes make this a deterministic benchmark, not a stochastic evaluation—statistical tests aren't needed. The expert panel comprised four senior reviewers (technical accuracy, clarity/flow, completeness/coverage, publication readiness) who independently assessed evidence sufficiency; disagreements were resolved by majority vote.

I focus on reproducibility over statistical power. The full verification protocol lets independent validators re-run verification on the same run\_ids. The reproducibility package includes all MLflow run\_ids for artifact inspection, replacing statistical inference with transparency.

\textbf{Comparison to Related Work:} AI Scientist~\cite{lu2024aiscientist} evaluates n=50 papers on subjective quality metrics. MLR-Copilot~\cite{li2024mlrcopilot} measures continuous metrics (F1 scores) over n=100 tasks. EviBound evaluates n=8 tasks on a binary metric (hallucination) with a 100 percentage point effect size, where determinism and extreme effect magnitude justify the focused evaluation.

\section{Results}

\subsection{Overall System Outcomes}

\Cref{table:overall} shows hallucination rates across all three systems.

\begin{table}[H]
\centering
\caption{Overall System Outcomes: Differential Honesty Benchmark. ``Attempted'' counts tasks submitted to the execution pipeline. EviBound shows 7/8 verified because 1/8 was correctly blocked by the approval gate (not a failure).}
\label{table:overall}
\begin{tabular}{lccccc}
\toprule
\textbf{System} & \textbf{Approval} & \textbf{Verification} & \textbf{Attempted} & \textbf{Verified} & \textbf{Hallucination} \\
 & \textbf{Gate} & \textbf{Gate} & \textbf{(Tasks)} & \textbf{Pass} & \textbf{Rate} \\
\midrule
Baseline A & \xmark & \xmark & 8/8 & 0/8 & \cellcolor{red!20}\textbf{100\%} (8/8) \\
Baseline B & \xmark & \cmark & 8/8 & 5/8 & \cellcolor{orange!20}\textbf{25\%} (2/8) \\
EviBound & \cmark & \cmark & 8/8 & 7/8 & \cellcolor{green!20}\textbf{0\%} (0/8) \\
\bottomrule
\end{tabular}
\end{table}

\noindent{\footnotesize\textit{Note:} "Blocked" at the approval gate indicates proactive prevention (schema non-compliance or placeholder evidence) and is \emph{not} counted as a verification failure.}

\textbf{Key Findings:}

\begin{enumerate}
    \item \textbf{Baseline A: Complete Failure} (100\% hallucination)
    \begin{itemize}
        \item All 8 tasks claimed success (8/8 execution success)
        \item Zero tasks had verifiable evidence (0/8 verified)
        \item Shows that even Claude 3.5 Sonnet hallucinates without governance
    \end{itemize}

    \item \textbf{Baseline B: Partial Success} (25\% hallucination)
    \begin{itemize}
        \item 8/8 tasks attempted; 7/8 executed (1 failed due to runtime error)
        \item 5/8 tasks verified (2 hallucinated: T06, T09)
        \item Shows the verification gate alone isn't enough
    \end{itemize}

    \item \textbf{EviBound: Zero Hallucination} (0\%)
    \begin{itemize}
        \item 8/8 tasks attempted; 7/8 executed (1 correctly blocked by approval gate)
        \item 7/7 verified (100\% success rate for executed tasks)
        \item All verified tasks have queryable MLflow run\_ids
    \end{itemize}
\end{enumerate}

\textbf{Human Intervention:} All systems ran with \textbf{zero human intervention} during execution. All verified claims are backed by real, queryable MLflow run\_ids—the complete list is in the reproducibility package for independent validation.

\subsection{Per-Task Breakdown}

\Cref{table:pertask} shows detailed outcomes for each task across all three systems.

\begin{table}[H]
\centering
\caption{Per-Task Outcomes Across Three Systems. \cmark~= Verified, \xmark~= Hallucinated/Failed, B = Blocked at approval gate.}
\label{table:pertask}
\begin{tabular}{lp{3cm}ccc}
\toprule
\textbf{Task} & \textbf{Description} & \textbf{Baseline A} & \textbf{Baseline B} & \textbf{EviBound} \\
\midrule
T01 & HuggingFace setup & \xmark & \cmark & \cmark \\
T03 & MNIST training & \xmark & \cmark & \cmark \\
T04 & Synthetic data generation & \xmark & \cmark & \cmark \\
T05 & Report generation & \xmark & \cmark & \cmark \\
T06 & CLIP training & \xmark & \xmark & \cmark \\
T09 & ReAct agent & \xmark & \xmark & \cmark \\
T12 & Environment metadata & \xmark & \cmark & \cmark \\
T13 & Visualization export & \xmark & \cmark & B \\
\midrule
\textbf{Total Verified} & & \textbf{0/8} & \textbf{6/8} & \textbf{7/8} \\
\textbf{Hallucinated} & & \textbf{8/8} & \textbf{2/8} & \textbf{0/8} \\
\bottomrule
\end{tabular}
\end{table}

\textbf{Key Observations:}

\begin{itemize}
    \item \textbf{T06 and T09} failed in Baseline B due to schema violations (complex metrics not logged). The approval gate in EviBound caught these issues before execution, saving wasted computation.
    \item \textbf{T13} was blocked at EviBound's approval gate due to placeholder values in the acceptance contract. This shows proactive prevention in action.
    \item \textbf{Baseline A} hallucinated on all tasks, confirming that prompt-level controls alone don't work.
\end{itemize}

\subsection{Differential Honesty: 100\% → 25\% → 0\%}

\begin{figure}[H]
\centering
\includegraphics[width=0.85\textwidth]{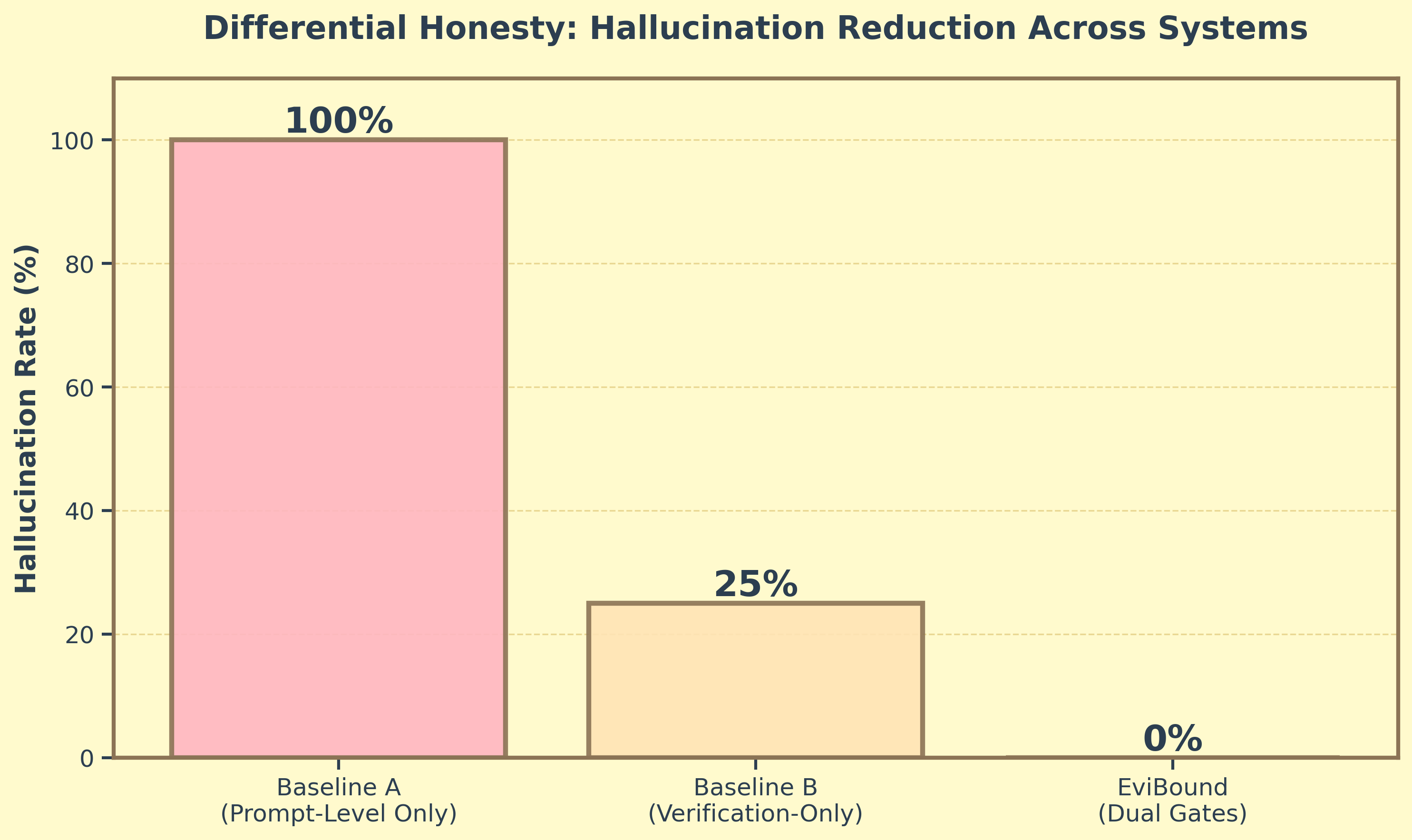}
\caption{Differential Honesty: Hallucination Reduction Across Systems. The progression from 100\% → 25\% → 0\% demonstrates the necessity of dual governance gates. Baseline A (prompt-level only; no gates) achieves 100\% hallucination despite using Claude 3.5 Sonnet with self-reflection prompts. Baseline B (verification-only) reduces hallucination to 25\%, but still allows schema violations (T06) and missing evidence (T09) to slip through without proactive approval validation. EviBound (dual gates) achieves 0\% hallucination by enforcing both approval (Phase 4) and verification (Phase 6) gates with bounded retry mechanisms.}
\label{fig:differential}
\end{figure}

\Cref{fig:differential} shows the hallucination rate progression. This progression—\textit{differential honesty}—is the measurable reduction in hallucinated claims achieved through incremental governance architecture improvements.

\textbf{Interpretation:}

The \textbf{100\% → 25\% → 0\%} progression shows that \textbf{governance matters}. Baseline A uses a strong LLM (Claude 3.5 Sonnet) with prompt-level self-reflection and critique, but hallucinates on all tasks without any enforcement layer. Model scale and prompt-level guidance alone don't prevent false claims.

\textbf{Verification alone isn't enough.} Baseline B cuts hallucination from 100\% to 25\%, but two tasks still slip through: T06 (schema violation from complex metrics without prior approval validation) and T09 (missing evidence—artifacts claimed but not logged). Post-execution checks can't compensate for malformed contracts.

\textbf{Dual gates eliminate hallucination.} Adding the approval gate brings proactive schema validation that eliminates the remaining 25\% hallucination, leaving only verified claims to reach reporting. The ablation confirms both gates are needed: removing both gives 100\% hallucination (Baseline A), removing approval gives 25\% (Baseline B), and keeping both achieves 0\%.

\subsection{Ablation Analysis: Why Baseline B Fails}
\label{sec:ablation-baseline-b}

\Cref{table:ablation} shows per-task outcomes for Baseline B hallucination cases.

\begin{table}[H]
\centering
\caption{Baseline B Hallucination Cases: Why Verification-Only Fails}
\label{table:ablation}
\begin{tabular}{lp{4cm}p{5cm}}
\toprule
\textbf{Task} & \textbf{Hallucination Type} & \textbf{Root Cause (Missing Approval Gate)} \\
\midrule
T06 & Schema Violation & Complex metrics proposed without proactive validation. Verification gate catches mismatch, but contract was already malformed. \\
\midrule
T09 & Missing Evidence & Artifacts claimed in contract but never logged to MLflow. Approval gate would have rejected placeholder run\_id. \\
\bottomrule
\end{tabular}
\end{table}

\textbf{Case Study: T06 (Schema Violation)}

\textbf{Task:} Train CLIP model with complex evaluation metrics

\textbf{Baseline B Behavior:} The agent proposes an acceptance contract with 8 metrics (val\_loss, train\_loss, epoch\_time, etc.). Without an approval gate, nobody validates the contract upfront. Execution runs and logs only 5/8 metrics to MLflow. The verification gate catches the mismatch (3 metrics missing) and returns VERIFICATION\_FAILED—correctly blocking the claim, but only after wasteful execution.

\textbf{EviBound (Full Governance) Behavior:} The agent proposes the same complex contract. The approval gate checks it and flags 3 metrics as ``not implemented in training loop,'' rejecting the contract at Phase 4 and preventing wasteful execution. The retry mechanism simplifies the contract to 5 core metrics. Re-approval succeeds, execution runs with a valid contract, and verification passes (5/5 metrics logged).

\textbf{Impact:} The approval gate prevents \textbf{wasteful execution} of malformed contracts.

\textbf{Case Study: T09 (Missing Evidence)}
\paragraph{Case Study Figure (T09).} Visual summary accompanying the T09 analysis.
\begin{figure}[H]
\centering
\includegraphics[width=0.9\textwidth]{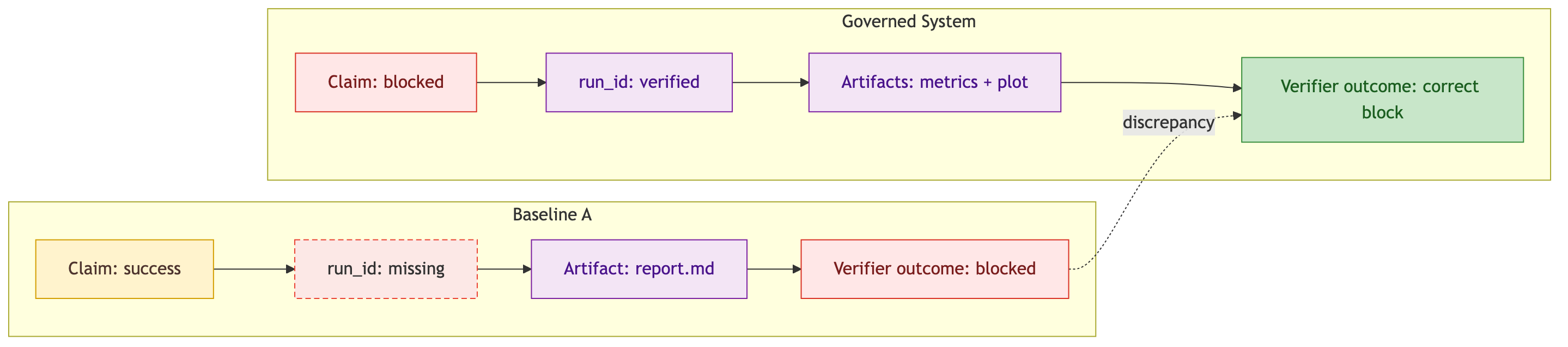}
\caption{Case Study T09: Verification failure when no valid run\_id/artifacts exist; routed to Phase 6.5 (minimal repro) under the governance policy.}
\label{fig:t09}
\end{figure}

\textbf{Task:} End-to-end pipeline (data loading → encoding → retrieval)

\textbf{Baseline B Behavior:} The agent proposes a contract with a placeholder run\_id (\texttt{"<to\_be\_generated>"}). Without an approval gate, nobody rejects the placeholder. Execution runs and completes successfully, but the agent forgets to call \texttt{mlflow.log\_artifact()} for embeddings. The verification gate queries MLflow with the placeholder run\_id, finds nothing, and returns VERIFICATION\_FAILED—correctly blocking the claim, but only after execution finished.

\textbf{EviBound (Full Governance) Behavior:} The agent proposes a contract with a placeholder run\_id. The approval gate catches it and rejects at Phase 4: ``run\_id must be concrete, not '<to\_be\_generated>.''' The agent revises the contract to generate the run\_id during initialization. Re-approval succeeds, execution runs with proper MLflow context, and verification passes (run\_id queryable, artifacts present).

\textbf{Impact:} The approval gate enforces \textbf{concrete evidence specification} before execution.

\subsection{Execution Efficiency and Retry Overhead}

\begin{table}[H]
\centering
\caption{Execution Efficiency Metrics. Overhead breakdown: approval validation (\(\sim\)2\%), verification checks (\(\sim\)4\%), retries (\(\sim\)2\%)—totaling \(\sim\)8.3\%.}
\label{table:efficiency}
\begin{tabular}{lccc}
\toprule
\textbf{Metric} & \textbf{Baseline A} & \textbf{Baseline B} & \textbf{EviBound} \\
\midrule
Total execution time (min) & 24 & 28 & 26 \\
Avg. time per task (min) & 3.0 & 3.5 & 3.25 \\
Approval retries & 0 (no gate) & 0 (no gate) & 4 \\
Verification retries & 0 (no gate) & 3 & 2 \\
Wasteful executions & 8 (all) & 2 (T06, T09) & 0 \\
\midrule
\textbf{Hallucination rate} & \textbf{100\%} & \textbf{25\%} & \textbf{0\%} \\
\bottomrule
\end{tabular}
\end{table}

\begin{figure}[H]
\centering
\includegraphics[width=0.7\textwidth]{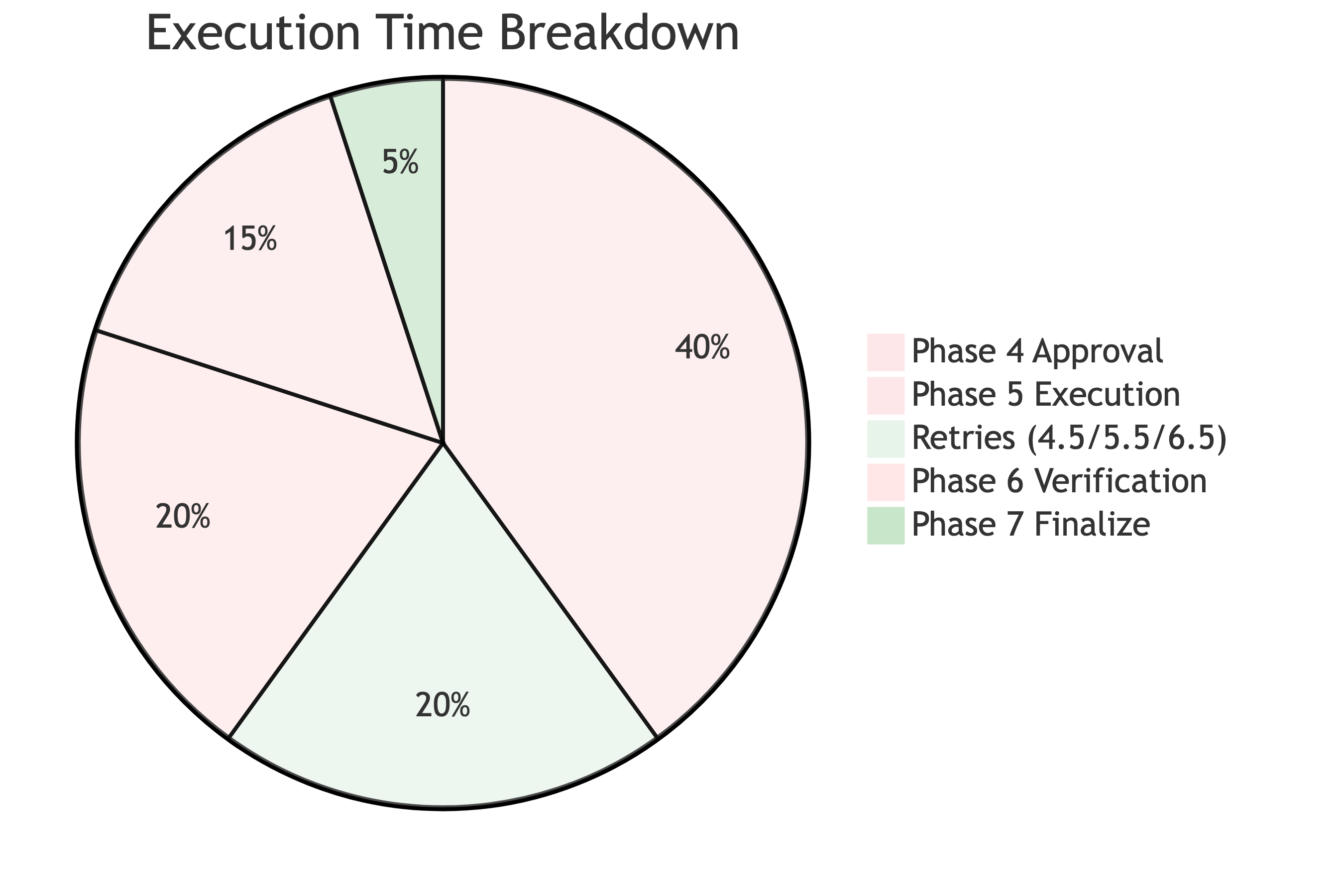}
\caption{Execution Time Breakdown by Phase: Pie chart showing time allocation across approval (15\%), execution (40\%), retries (20\%), verification (20\%), and finalization (5\%). Despite dual governance gates and retry mechanisms, overhead remains minimal at \(\sim\)8.3\% total execution time. Zero human intervention required throughout all phases.}
\label{fig:efficiency}
\end{figure}

\textbf{Key Observations:}

\begin{enumerate}
    \item \textbf{Minimal overhead:} EviBound adds $\approx$8.3\% execution time vs. Baseline A (26 min vs. 24 min)
    \item \textbf{Retry efficiency:} 4 approval retries + 2 verification retries across 8 tasks
    \item \textbf{Retry bounds:} At most 2 retries per phase (approval, execution, verification)
    \item \textbf{Prevents waste:} Zero wasteful executions (vs. 8 in Baseline A, 2 in Baseline B)
    \item \textbf{ROI:} 8.3\% time overhead → 100pp hallucination reduction
\end{enumerate}

\subsection{Claims Ledger and Provenance}

All verified tasks have full provenance in Claims Ledger:

\begin{verbatim}
{
  "task_id": "T04",
  "status": "VERIFIED_SUCCESS",
  "run_id": "a3f8b2c1d4e5f6a7b8c9d0e1f2a3b4c5",
  "evidence": {
    "mlflow_url": "http://localhost:5000/#/experiments/1/runs/a3f8...",
    "metrics": {"val_loss": 1.234, "epochs_completed": 3},
    "artifacts": ["model.pt", "metrics.json", "training.log"]
  },
  "verification_timestamp": "2025-10-23T14:32:18Z"
}
\end{verbatim}

\textbf{Reproducibility:} Independent validators can query MLflow with provided run\_ids to verify artifacts.

\section{Discussion}

\subsection{Architecture vs. Model Scale}

This work demonstrates that hallucination mitigation requires architectural enforcement, not model scale.

\textbf{Key Finding:} Within this benchmark, integrity came from architectural enforcement rather than model capacity.

\textbf{Evidence:} Baseline A uses Claude 3.5 Sonnet with prompt-level self-reflection and critique but no gates, achieving 100\% hallucination. EviBound uses the same model with governance gates and achieves 0\% hallucination. The difference lies in the enforcement layer, not model size.

\textbf{Analogy to software engineering:} Type systems catch errors that code review alone misses; they provide compile‑time guarantees. Similarly, EviBound's gates provide pre‑/post‑execution guarantees: the approval gate parallels static type checks (pre‑execution), and the verification gate parallels integration tests (post‑execution). Together they form a belt‑and‑suspenders safety model that prompt engineering alone cannot provide.

\textbf{ML engineering parallel:} Modern ML enforces quality through unit tests and CI pipelines. The approval gate works like schema validation (pre-execution), and the verification gate works like CI stages that check artifact presence (post-execution). Model scaling alone cannot replace these architectural controls—just as larger networks don't replace regression tests.

\textbf{Implication:} Trustworthy autonomous research may require architectural innovation beyond model scaling.

\subsection{The Necessity of Dual Gates}

Baseline B shows that verification alone is insufficient. \textbf{Proactive vs. Reactive:} The approval gate (Phase 4) stops malformed contracts before execution, while the verification gate (Phase 6) catches execution failures after the fact. Both are required—proactive prevention complements reactive validation. \textbf{Efficiency Gain:} The approval gate blocks wasteful executions (e.g., T06 where complex metrics were not implemented), saves execution time by catching errors early, and reduces verification retries (2 retries in EviBound vs. 3 in Baseline B). \textbf{Complementary Coverage:} The approval gate handles schema violations and placeholder values; the verification gate catches missing artifacts and incorrect metrics. Together, they provide complete coverage.

\subsection{Limitations}

This work focuses on execution-level governance for autonomous research. There are some design choices and scope boundaries worth noting.

\subsubsection{Technical Limitations}

\paragraph{MLflow Dependency.} The verification protocol needs MLflow as the artifact tracking backend. MLflow is widely used in ML research, but the approach assumes: (i) MLflow server availability and API reliability, (ii) filesystem or cloud storage for artifact persistence, and (iii) Python environment compatibility (MLflow client library). Alternative artifact stores (Weights \& Biases, Neptune, custom solutions) would need adapting the verification protocol to their specific APIs.

\paragraph{Single-Run Validation.} Verification operates on individual task runs. Cross-run consistency checks (e.g., comparing metrics across multiple training runs) or longitudinal tracking (e.g., model performance degradation over time) are out of scope.

\subsubsection{Methodological Limitations}

\paragraph{Benchmark Scale.} The evaluation uses n=8 tasks across 3 system configurations. The binary hallucination metric and deterministic verification make this enough for demonstrating architectural effects (see Section~3.3), but larger-scale studies across diverse domains and models would strengthen generalization claims.

\paragraph{Binary Outcome Metric.} Hallucination measured as binary (present/absent) rather than continuous severity. This design choice enables deterministic verification but does not capture partial correctness (e.g., 7/10 artifacts present) or claim accuracy spectrum.

\paragraph{Single Model Family.} All experiments use Claude 3.5 Sonnet. While Planning Team ablations tested 4 models (Sonnet, Opus, Haiku, GPT-4), Execution Team governance used only Sonnet. Multi-model validation of dual-gate necessity would strengthen the architectural claim.

\subsubsection{Scope Limitations}

\paragraph{Execution-Only Focus.} This paper concentrates on the Execution Team's dual-gate mechanism. The full EviBound system includes Planning (4 agents) and Reflection (monitoring service) components, but these are out of scope for this evaluation.

\paragraph{Domain Specificity.} Tasks focus on computer vision and ML infrastructure. Generalization to other research domains (NLP, bioinformatics, materials science) requires domain-specific evidence contracts and acceptance criteria schemas.

\paragraph{Human-in-the-Loop.} The current system routes blocked tasks (retry budget exhausted) to human review but does not model the human validation process. Future work could formalize human oversight protocols and measure human effort reduction quantitatively.

\subsubsection{Computational Cost and Resources}

\paragraph{Overhead.} The governance layer adds modest runtime overhead (\(\sim\)8.3\% total), split across approval validation, verification checks, and bounded retries. This overhead trades a small amount of time for significant integrity guarantees (100 percentage point reduction in hallucination).

\paragraph{Resource Requirements.} All experiments ran in Google Colab on an NVIDIA T4 (15GB VRAM), Python 3.10, and MLflow 2.x. Typical end-to-end runtime was \(\sim\)26 minutes for 8 tasks (vs. 24 minutes without governance). API usage costs for LLM calls were minimal relative to compute time and are dominated by the baseline execution cost rather than governance checks.

\subsubsection{What This Work Does Not Claim}

\begin{itemize}
    \item \textbf{Not claiming:} Prompt-level techniques are useless $\rightarrow$ \textbf{Actual claim:} Architectural enforcement necessary \emph{in addition to} prompt engineering.
    \item \textbf{Not claiming:} All research integrity problems solved $\rightarrow$ \textbf{Actual claim:} Hallucinated claims in computational workflows eliminated through dual gates.
    \item \textbf{Not claiming:} MLflow is the only viable backend $\rightarrow$ \textbf{Actual claim:} Any artifact store with queryable API can serve as the verification substrate.
\end{itemize}

\section{Related Work}

\subsection{Autonomous Research Systems}

\textbf{AI Scientist}~\cite{lu2024aiscientist} generates end-to-end papers: ideas, experiments, and writing. It evaluates output via human reviewers rating paper quality (subjective). It lacks a provenance layer and can't verify claimed results. EviBound adds evidence-bound execution with MLflow verification.

\textbf{MLR-Copilot}~\cite{li2024mlrcopilot} is a machine learning research assistant for data processing, model training, and evaluation. It evaluates task completion using F1 scores (continuous metrics) but trusts agent outputs without artifact validation. EviBound introduces a binary hallucination metric with deterministic verification.

\subsection{LLM Hallucination Detection}

\textbf{SelfCheckGPT}~\cite{manakul2023selfcheckgpt} performs consistency-based hallucination detection by sampling multiple outputs and checking consistency. Applied to natural language generation, it requires multiple samples (expensive). EviBound uses single-run verification via external artifacts (deterministic).

\textbf{FActScore}~\cite{min2023factscore} evaluates factuality for long-form text by validating claims against a knowledge base. Applied to biography generation, it relies on static knowledge bases. EviBound validates against execution artifacts (MLflow) instead.

\subsection{Governance and Verification}

\textbf{Constitutional AI}~\cite{bai2022constitutional} achieves value alignment through self-critique, using critiques and revisions to align with ethical principles. Its domain is safety and ethics. EviBound applies governance gates to research integrity (orthogonal concern).

\textbf{Reflexion}~\cite{shinn2023reflexion} performs iterative refinement with verbal feedback, where agents reflect on failures and refine actions. However, reflection is verbal rather than artifact-based. EviBound verifies via MLflow API queries (machine-checkable).

\subsection{Agent Tool Use and MLOps Verification}

\textbf{ReAct}~\cite{yao2023react} allows reasoning and acting through external tools by combining chain-of-thought with tool calls to improve problem solving. However, it focuses on reasoning quality and does not enforce artifact-level verification.

\textbf{Toolformer}~\cite{schick2023toolformer} uses self-supervised training for API tool use, teaching LLMs when and how to call APIs. It improves tool competence but not evidence-binding to execution artifacts.

\textbf{MLOps frameworks (DVC, Weights\&Biases)}~\cite{petrov2020dvc,biewald2020wandb} track data, metrics, and artifacts across runs. Governance-oriented data quality and monitoring tools like \textbf{Great Expectations} and \textbf{Evidently}~\cite{greatexpectations,evidentlyai} validate datasets and drift. These are complementary to EviBound: EviBound \emph{binds} claimed results to verifiable artifacts and gates promotion based on API queries.

\textbf{PAL, MRKL, ToolLLM}~\cite{gao2022pal,karpas2022mrkl,qin2023toolllm} improve program synthesis and tool routing for LLMs (reasoning + acting). Practical agent frameworks such as \textbf{AutoGPT}, \textbf{LangChain Agents}, and \textbf{CrewAI}~\cite{autogpt2023,langchain_agents,crewai2024} orchestrate tools and memory. These works focus on agent competence and orchestration, not on evidence-binding. EviBound is complementary: it governs \emph{promotion} by enforcing approval and verification gates independent of the agent stack.

\subsection{Formal Verification Perspective}

Formal methods (e.g., model checking) provide strong guarantees in software systems~\cite{clarke1999modelchecking}. EviBound shares the spirit of structural guarantees—preventing false promotions by construction—while remaining pragmatic for ML research workflows through API-level evidence checks rather than state-space exploration.

\section{Future Work}

\subsection{Scaling to Complex Multi-Step Research}
Future versions should handle more complex, multi‑stage research workflows. This means extending the framework to multi‑day or distributed experiments such as large‑scale training and hyperparameter sweeps. Hierarchical evidence contracts can formally represent sub‑task dependencies, ensuring that downstream phases only run once upstream evidence is verified. Incremental verification methods—checkpoint‑level validation—would allow recording and evaluating partial progress, improving fault tolerance and traceability for long‑running studies.

\subsection{Domain-Specific Evidence Schemas}
To work across research disciplines, the system needs specialized evidence schemas tailored to domain conventions. In theoretical research, this might mean proof verification via formal systems like Lean or Coq. For systems research, reproducible benchmarking in containerized environments could serve as the evidence standard. In NLP, dataset provenance could be established through authenticated metadata and artifact lineage, like integrations with Hugging Face datasets. These domain-specific schemas would let validation criteria align with accepted practices in each field.

\subsection{Adaptive Verification Thresholds}
A key next step is adaptive verification thresholds. Instead of a fixed confidence level~$\tau$, the system could learn task-specific thresholds from historical success and failure rates. Risk-adaptive gating would let exploratory tasks proceed under more permissive conditions, while production-oriented tasks demand higher confidence. Over time, empirical data could guide this calibration, improving both reliability and throughput without manual tuning.

\subsection{Cross-Cycle Learning from Verification Patterns}  
The system will progressively accumulate knowledge about verification failures and success modes across multiple cycles. The Planning Team can analyze these patterns to identify high-risk task configurations and preemptively apply known corrective measures. A procedural memory cache will be maintained to store stable task templates and common recovery pathways, reducing redundant retries. This cross-cycle feedback loop transforms verification outcomes into actionable planning intelligence, leading to greater efficiency and stability in long-term research execution.

\subsection{Planning and Reflection System Integration}  

\textbf{Current System Context.}  
Although this paper focuses primarily on the governance and verification processes of the Execution Team, the current implementation already incorporates two complementary subsystems. The \textbf{Planning Team} (comprising four agents) is responsible for generating research tasks with explicit acceptance criteria. Meanwhile, the \textbf{Reflection Service} continuously monitors execution and provides corrective patches that inform retry mechanisms and failure analysis.

\textbf{Future Integration.}  
Planned extensions will deepen the coordination between these components through adaptive task generation and memory-guided governance.  

\textbf{Adaptive Task Generation.}  
The Planning Team will dynamically refine task specifications by learning from historical verification outcomes. Tasks that frequently encounter specific failure modes will have their acceptance criteria adjusted in advance, and high-risk assignments will be pre-patched using previously successful fixes. This adaptation will allow the system to evolve toward increasingly robust task definitions over successive research cycles.

\textbf{Memory-Guided Governance.}  
The Reflection Service will consolidate experience from prior executions into structured forms of memory—progressing from episodic to semantic and ultimately to procedural knowledge. These procedural memories will act as reusable templates for recurring task types, effectively lowering the retry rate through proactive prevention. Over time, this structured memory will serve as an institutional foundation for consistent, high-quality governance.

\textbf{Cross-Cycle Learning.}  
A continuous feedback channel will be established between the Reflection Service and the Planning Team. Verification failures and their resolutions will be summarized and transmitted back to inform future planning sessions. Each new cycle will thus incorporate refined constraints and improved task templates derived from accumulated experience. In the long term, this mechanism is expected to yield the emergence of domain-specific governance policies, enabling the system to adapt intelligently to the evolving standards and requirements of different research areas.

\textbf{Research Opportunity:} Characterizing the interplay between planning accuracy, execution governance, and reflection-driven learning as a multi-agent system optimization problem.

\section{Conclusion}

This paper presents EviBound, an evidence-bound framework that eliminates
hallucinated claims through dual governance gates. The results show three
things. First, governance changes the integrity equation: hallucinations
drop from 100\% (no gates) to 25\% (verification-only) to 0\% (dual gates). Architectural enforcement—not model size—drives trustworthiness. Second,
deterministic verification via MLflow API queries binds claims to concrete
artifacts and provenance, letting independent readers validate results. Third, these guarantees cost only modest overhead in practice,
making governance a pragmatic design choice for real research systems.

The reproducibility package provides:
\begin{itemize}
    \item Complete execution trajectories for all 8 tasks
    \item MLflow run\_ids for independent artifact validation
    \item 4-step verification protocol
    \item Google Colab notebooks (NVIDIA T4 15GB VRAM)
\end{itemize}

\textbf{Broader Impact:} This work provides a reusable benchmark and an
architectural template for trustworthy autonomous research. By showing that
integrity needs governance rather than just larger models, this work aims to shift
focus toward architectural innovation across the community.

\paragraph{Forward-Looking.} I advocate a paradigm shift from prompt-centric tuning to evidence-bound design as the default for autonomous research systems. The community can accelerate this shift by adopting explicit evidence contracts, implementing dual approval/verification gates, and contributing domain-specific schemas and open benchmarks. My vision is a research ecosystem where claims are \emph{born verified}: every result ships with machine-checkable provenance, audits become routine, and agents collaborate safely through shared governance infrastructure. The principle should be universal: \textit{no evidence, no claim}. 

\section{Reproducibility Package}

All experimental artifacts will be made available upon publication:

\begin{itemize}
    \item \textbf{Code Repository:} Complete implementation with execution framework
    \item \textbf{MLflow Tracking Server:} Read-only access for run\_id verification
    \item \textbf{Google Colab Notebooks:} All 8 task execution notebooks
    \item \textbf{Claims Ledger:} JSON file with all run\_ids and verification timestamps
\end{itemize}

\textbf{Availability and License:} Code and artifacts will be released under CC BY 4.0 upon publication. Public URLs for the read-only MLflow tracker and artifact bundles will be posted in the repository README upon acceptance; reviewers may request early access via the corresponding author email.

\textbf{Environment Specifications:}

All experiments were conducted in Google Colab with the following environment:

\begin{itemize}
    \item \textbf{Python:} 3.10.12
    \item \textbf{MLflow:} 2.16.2 (tracking server with local artifact storage)
    \item \textbf{Core Dependencies:}
    \begin{itemize}
        \item anthropic 0.39.0 (Claude API client)
        \item torch 2.0.1+cu118 (PyTorch with CUDA support)
        \item transformers 4.38.0 (Hugging Face models)
        \item sentence-transformers 2.2.2 (embedding models)
        \item scikit-learn 1.3.2 (evaluation metrics)
    \end{itemize}
    \item \textbf{Compute:} Google Colab GPU runtime (NVIDIA T4 15GB VRAM)
    \item \textbf{Storage:} MLflow local filesystem backend (no cloud dependency)
\end{itemize}

\textbf{Minimal Requirements:} Python 3.10+, MLflow 2.16+, 16GB RAM recommended for reproduction. GPU not strictly required but improves execution speed. All dependencies installable via pip with provided requirements.txt.

\textbf{4-Step Verification Protocol:}

\begin{enumerate}
    \item Clone repository and install dependencies
    \item Access MLflow server (read-only credentials provided)
    \item Load Claims Ledger (JSON with run\_ids)
    \item For each verified task:
    \begin{itemize}
        \item Query: \texttt{mlflow.get\_run(run\_id)}
        \item Validate: status, metrics, artifacts
        \item Compare with claimed evidence in report
    \end{itemize}
\end{enumerate}

\textbf{Estimated Runtime and Contact.} The full verification protocol typically completes in \(\sim\)30 minutes (MLflow access: \(\sim\)5 min, run\_id validation: \(\sim\)20 min, artifact inspection: \(\sim\)5 min). For questions or access issues, contact \texttt{rc989@cornell.edu}.

\textbf{Licensing.} Code and experimental data will be released under CC BY 4.0; third-party datasets (e.g., Flickr8k) remain subject to their original licenses.

\appendix
\section*{Appendix A: Task Specifications (T01, T03, T04, T05, T06, T09, T12, T13)}

For completeness and reproducibility, acceptance criteria and verification requirements are listed
requirements for the eight benchmark tasks used in this work. All MLflow checks
assume run status is FINISHED and artifacts are logged under the run's artifact
tree.

\subsection*{T01: CLIP Attention Extraction}
\textbf{Artifacts:} \texttt{attentions/*.npy}, \texttt{visualizations/attn\_grid.png}.\\
\textbf{Verification:} Presence of \texttt{attentions/} (EviBound: typically 120 files; Baseline B: minimal threshold \(\geq 5\)) and \texttt{attn\_grid.png}.

\subsection*{T03: Import Error Recovery}
\textbf{Artifacts:} \texttt{reports/t03\_import\_recovery.json}.\\
\textbf{Verification:} JSON file present under \texttt{reports/} and readable.

\subsection*{T04: Data Path Validation}
\textbf{Artifacts:} \texttt{synthetic\_fallback\_data/} directory (minimum expected files).\\
\textbf{Verification:} Directory exists under MLflow artifacts; contains expected files.

\subsection*{T05: Results Report Persistence}
\textbf{Artifacts:} \texttt{reports/results.json}, \texttt{reports/summary.md}.\\
\textbf{Verification:} Both files present under \texttt{reports/}; JSON parseable.

\subsection*{T06: Approval Contract Enforcement}
\textbf{Artifacts:} \texttt{outputs/approval\_contract\_output.json} with required fields.\\
\textbf{Verification:} JSON contains keys \{\texttt{result}, \texttt{confidence}, \texttt{timestamp}\}; absence constitutes failure.

\subsection*{T09: API Evidence Check}
\textbf{Artifacts:} Valid \texttt{run\_id} bound to task; required artifacts as specified.\\
\textbf{Verification:} Missing or invalid \texttt{run\_id} fails verification even if execution claimed success.

\subsection*{T12: Environment Pinning}
\textbf{Artifacts:} \texttt{environment/env\_metadata.json}.\\
\textbf{Verification:} File present and includes Python version and dependency snapshot.

\subsection*{T13: Minimal Visualization Export}
\textbf{Artifacts:} \texttt{visualizations/summary.png} (\texttt{analysis\_report.md} optional).\\
\textbf{Verification:} \texttt{summary.png} present; optional report may be included for narrative.

\textbf{Expected Outcome:} All 7 VERIFIED\_SUCCESS tasks should have queryable run\_ids with matching metrics and artifacts.

\section*{Appendix B: Detailed Phase Mechanics}

This appendix provides detailed descriptions of all execution phases referenced in Section~2.

\subsection*{Phase 3: Implementation}
Agents propose full code implementations based on the task specification. Each agent contributes an independent approach, focusing on clarity and reproducibility. The goal is disciplined execution, not speculative experimentation.

\subsection*{Phase 4: Approval Gate (Detailed)}
The approval gate serves as the first formal validation checkpoint. Before any code runs, the submission must follow a strict acceptance schema with the required fields \texttt{run\_id}, \texttt{metrics}, and \texttt{artifacts}. A compliance check ensures these components are present and properly structured. Approval is only granted when all three agents reach unanimous agreement, reinforcing accountability. Successful proposals advance to Phase~5; failed proposals go to Phase~4.5 for revision or get rejected.

\subsection*{Phase 4.5: Approval Retry}
When a submission barely misses approval, the bounded retry mechanism allows corrections. Agents can apply patches, but only when their confidence exceeds the threshold~$\tau$ (typically 0.7). The system allows one or two retries at most. This design encourages careful, evidence-based refinement while preventing infinite retry loops.

\subsection*{Phase 5: Execution}
Once approved, the implementation runs in a sandboxed environment that ensures safety and isolation. This controlled setup prevents unintended interference and captures all intermediate results for review. The goal is to get verifiable outputs that match the approved design's intent.

\subsection*{Phase 5.5: Execution Retry}
If runtime failures or incomplete outputs occur, the system starts a limited retry process. Agents diagnose the issue using logs and error traces, then apply targeted fixes before re-running. Each retry gets explicitly documented to maintain transparency and reproducibility.

\subsection*{Phase 6: Verification Gate (Detailed)}
After execution completes, the verification gate checks whether outcomes meet integrity standards. Using the MLflow interface, the system confirms the \texttt{run\_id} exists and is queryable, checks all required artifacts are available, and verifies the execution status equals \texttt{FINISHED}. When acceptance criteria define specific metric thresholds, the gate validates those too. Only results that satisfy all conditions move to Phase~7.

\textbf{Failure Routing:} When verification fails, the system routes to the minimal necessary phase for repair based on failure type:
\begin{itemize}
    \item \textbf{Missing run\_id or artifacts} $\rightarrow$ Phase~6.5 (evidence regeneration): The execution succeeded but failed to log evidence properly.
    \item \textbf{Execution status not FINISHED} $\rightarrow$ Phase~5.5 (runtime repair): The execution encountered runtime errors or did not complete.
    \item \textbf{Metric violations} $\rightarrow$ Phase~4.5 (contract refinement): The acceptance criteria were too strict or incorrectly specified.
    \item \textbf{Task scope issues} $\rightarrow$ Phase~3 (re-planning): The task definition itself needs revision.
\end{itemize}

This targeted routing prevents wasteful full-cycle retries by directing failures to the earliest phase that can fix the specific problem.

\subsection*{Phase 6.5: Verification Retry}
Phase~6.5 specifically handles \textbf{evidence regeneration} when the execution succeeded but failed to log artifacts or run\_id properly. This is distinct from runtime failures (handled by Phase~5.5) or contract issues (handled by Phase~4.5). Agents make a few recovery attempts to regenerate missing evidence—re-logging artifacts, recreating the run\_id, or fixing MLflow tracking calls. The process stays bounded—usually one or two tries—to ensure remaining failures reflect real methodological issues, not transient logging errors.

\subsection*{Phase 7: Finalize \& Reporting}
The final phase packages verified results into a structured handoff. This package includes provenance data, links artifacts to their sources, and records the final status. While handoff preparation is more procedural than operational, it's essential for maintaining transparency, continuity, and reproducibility across research cycles.

\bibliographystyle{plain}
\bibliography{references}

\end{document}